# A Rubric-Guided Large Language Model Solution for Opioid Use Disorder Computable Phenotyping

**Mengxian Lyu, MS[1], Paredes Pardo, MS[1], Cheng Peng, PhD[1], Ziyi Chen, MS[1], Mengyuan Zhang, MS[1], Jieting Li Lu, BS[2], Gary M Reisfield, M.D[3], William M Greene, M.D[3], Lo-Ciganic Jenny, PhD, MS, MSPharm[4*], Yonghui Wu, PhD[1, 5*]**

**[1]Department of Health Outcomes and Biomedical Informatics, College of Medicine, University of Florida, Gainesville, FL, USA; [2]Department of Engineering Education, Herbert Wertheim College of Engineering, University of Florida, Gainesville, FL, USA; [3]Department of Psychiatry, College of Medicine, University of Florida, Gainesville, FL, USA; [4]Division of General Internal Medicine, Department of Medicine, University of Pittsburgh School of Medicine, Pittsburgh, PA, USA; [5]Preston A. Wells, Jr. Center for Brain Tumor Therapy, Lillian S. Wells Department of Neurosurgery, University of Florida, Gainesville, Florida, USA.**

**Abstract**

*Opioid use disorder (OUD) remains a public health crisis in the United States, yet it is difficult to identify from electronic health records (EHRs) because missing diagnosis codes and supporting evidence are buried in clinical narratives. Accurate OUD identification is critical to support interventions and improve health outcomes. This study developed a rubric-guided large language model (LLM) that incorporated Optimization by PROmpting (OPRO) for OUD computable phenotyping (CP). The framework used an 18-item, expert-identified rubric to instruct LLMs to automatically extract critical text with supporting evidence to determine OUD flags. Two UF Health physicians (GMR and WMG) chart-reviewed 253 patients, including 68 OUD-positive cases. Our LLM-based computable phenotype (CP) achieved the best F1 score of 0.774 and an AUROC of 0.934, outperforming the machine learning-based CP using EHR and natural language processing-extracted variables, and zero-shot LLMs by relative F1 improvements of 12.8% and 44.4%, respectively. The proposed LLM-based CP could link LLM-extracted evidence to OUD phenotyping for better explainability.*

## Introduction

Opioid use disorder (OUD) is a major national public health threat in the United States[1,2]. The 2024 National Survey on Drug Use and Health estimated that 4.8 million people aged 12 years or older had OUD[3]. Healthcare systems propose various intervention programs to counter this crisis; however, such interventions require accurately identifying patients with OUD from electronic health records (EHRs), which is challenging. OUD is well-known as undercoded in EHRs; among patients prescribed opioids, it is estimated that only approximately 10% of them could be identified using OUD International Classification of Diseases (ICD) codes.[4] A study reported that across six U.S. primary-care systems, documented OUD diagnoses ranged from approximately 0.7% to 1.4% over three years[5]. Other evidence may be documented as heroin or fentanyl use, use of opioids other than as prescribed, medication treatment for OUD, injection-related complications, naloxone administration, polysubstance use, or opioid-seeking behavior[6]. Many of these cases may appear in the EHR without a clear OUD diagnosis. There is a critical gap in accurately identifying OUD cases from routinely collected EHR data to support clinical intervention and large-scale OUD studies.

In disease phenotyping, domain experts' chart review of EHRs is typically considered a gold standard. However, manual review is time-consuming and cannot scale up to support large-scale studies. Medical researchers typically develop a computable algorithm to automatically identify OUD cases by screening large-scale EHRs, which is known as a computable phenotyping (CP) algorithm. Previous studies have developed OUD CPs using rule-based or machine-learning models[7]. Rule-based CPs use domain experts to manually identify combinations of diagnosis codes, medications, laboratory results, and other EHR events. Rule-based CPs are efficient for screening large-scale EHRs, but they require substantial manual development and are difficult to expand or adapt across health systems[8]. Machine-learning-based CPs automatically learn important features from a larger number of structured EHR variables, including free-text variables that are extracted from clinical notes by natural language processing (NLP)[9]. Most CPs use structured EHRs, and a few explore clinical narratives, where NLP extracts information from clinical notes and encodes it into discrete features.

Large language models (LLMs) have become widely accepted as a standard solution for processing clinical narratives, offering a novel way to use whole-text clinical narratives instead of only small pieces of text extracted by NLP. Recent studies have applied LLMs to extract OUD-related information from clinical narratives and to identify patients with OUD from clinical notes[11,12]. LLMs can recognize OUD-related evidence and return structured evidence together with supporting excerpts from the source notes. These capabilities make LLMs particularly promising for evidence-based OUD computable phenotyping, in which clinically relevant observations can be identified from narrative documentation to further improve CP for OUD.

LLM-based phenotyping combines evidence identification and patient-level classification in a unified system [13]. This study aims to explore LLM-based CPs and develop algorithms to integrate evidence and rules previously identified in the literature. Two UF Health physicians chart-reviewed a cohort of 300 patients to create a gold standard dataset with 253 valid subjects. We explored LLM-based CPs and developed a method to integrate existing expert-identified evidence and rules to further improve CP for OUD. The framework comprised three components: OUD evidence items adapted from a physician-developed documentation ruleset, LLM-based extraction of item-level judgments with supporting text, and logistic regression for patient-level aggregation. We further used Optimization by PROmpting[14] (OPRO) to refine the evidence-extraction prompt and evaluated alternative evidence specifications and aggregation strategies. We compared our LLM-based solution with rule-based CP using diagnosis codes, machine learning-based CPs, and zero-shot LLM baselines. The chart-reviewed cohort is used to develop and evaluate the CP algorithms. The evaluation results show that our LLM-based CP achieved the best AUROC score of 0.934 and the best F1 score of 0.774, outperforming machine learning-based CPs by 12.8% and outperforming zero-shot LLMs by 44.4%. The proposed LLM-based CP improved accuracy in identifying OUD cases and provided better explainability by linking text evidence to CP determinations through the LLM's attention mechanism**.**

## Methods

### Overall study design

**Figure 1** shows an overview of our rule-guided LLM solution for OUD CP using structured EHR data and clinical narratives from UF Health. Physician chart-reviewed labels are used as the gold standard. The proposed LLM-based CP is composed of three components: an LLM engine (gpt-oss-120b[15]) that extracts evidence and determines the met/not-met binary judgments with supporting text evidence, and then a logistic regression model that uses the LLM-extracted indicators to generate patient-level OUD probabilities. The Optimization by PROmpting (OPRO) algorithm optimizes the evidence-extraction prompt. We compared our LLM-based CP with a diagnosis code–based phenotype, a machine learning-based CP using structured EHR features, NLP-derived features, and their combination, and a zero-shot LLM without rule guidance.

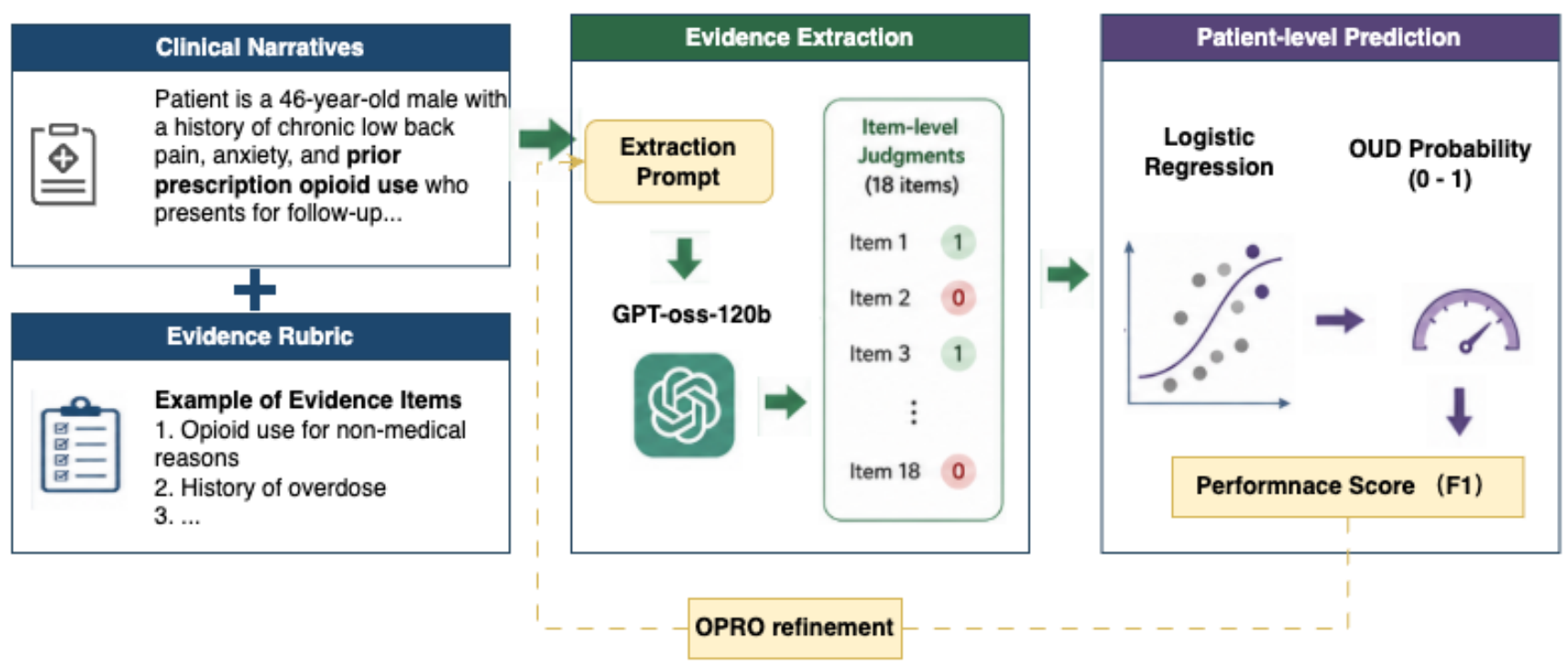

**Figure 1.** Rubric-guided LLM solution for OUD CP. An OUD evidence rubric adapted from an existing computable phenotype defines 18 evidence items for LLM extraction. Logistic regression combines the indicators to generate a patient-level OUD label.

### Cohort construction and chart review

We obtained patient-level structured EHR data and clinical notes from the University of Florida Health Integrated Data Repository under institutional review board approval (IRB202101897). We constructed an opioid-enriched candidate cohort by requiring subjects who have at least one of the following opioid-related signal, including (1) an opioid or overdose diagnosis, (2) an opioid or medication-for-OUD prescription, naloxone or methadone administration, or (3) an opioid laboratory result. Patients with 10–30 inpatient progress notes containing prespecified opioid-related terms were included. For each patient, an existing opioid-focused NLP system was used to select notes containing relevant clinical concepts for the review cohort. Two physicians (GMR and WMG) chart-reviewed 300 patients and assigned each patient an OUD label of TRUE, FALSE, or UNKNOWN. Both physicians reviewed the first 40 patients, and the remaining 260 patients were divided evenly between them after they achieved a good agreement score. Discordant labels were reconciled when possible; cases without a definitive classification were assigned UNKNOWN. TRUE and FALSE labels defined the positive and negative reference classes, respectively, whereas UNKNOWN labels were excluded from binary evaluation.

### Machine learning-based CPs

We developed machine learning-based CPs using three patient-level representations: structured EHR features alone, NLP-derived features alone, and combined EHR+NLP features. The structured EHR representation contained 1,508 diagnosis, medication, laboratory, utilization, demographic-category, and contextual variables. The NLP representation contained 237 features generated by an existing clinical NLP pipeline[11,15], including 126 clinical-concept features capturing opioid and OUD-related information such as opioid use, dependence, overdose, and related symptoms or interventions, and 111 social-determinants features covering substance use, employment, housing, transportation, and other social factors. Combining the two representations produced 1,745 EHR+NLP features. For each representation and outer training fold, 3-fold inner cross-validation selected the learner and hyperparameters from logistic regression, LASSO, elastic net, random forest, and XGBoost by average precision. All preprocessing, feature handling, model selection, and hyperparameter tuning were confined to the corresponding training data. For the matched comparison, a Youden threshold was estimated separately for each representation within each training fold and applied unchanged to the corresponding test fold. Thus, the representations shared the same threshold-selection rule rather than a single numeric threshold. As an additional fixed baseline, a patient was classified as positive by the diagnosis code baseline if at least one OUD diagnosis was recorded in the structured EHR.

### LLM-based CPs

**Zero-shot LLM classification**. We instruct LLMs to read patients' clinical narratives and return a patient-level OUD label, reason, and supporting text evidence. The prompt provided no task-specific examples and no itemized evidence rubric, which is known as a zero-shot setting. The model selected the relevant observations and produced the final judgment. This method served as an LLM baseline.

**Rubric-guided LLM CP**. The proposed LLM solution combined an OUD evidence rubric, LLM evidence extraction, patient-level classification, and an Optimization by PROmpting (OPRO)[14] module. To define the rubric, we adapted a physician-developed OUD documentation ruleset into 18 binary evidence items by separating confirmed, suspected, and other clinically distinct variants(**Table 1**)[16].

**Table 1.** The 18-item OUD evidence rubric used for LLM extraction.

| Item | Evidence definition |
|---|---|
| 1 | Fentanyl-positive urine test. |
| 2 | At least three IV drug-use-related complications. |
| 3a | Documented heroin use. |
| 3b | Treatment for heroin addiction. |
| 4a | Confirmed opioid misuse or abuse. |
| 4b | Suspected opioid misuse or abuse. |
| 5a | OUD or opioid addiction without active treatment. |
| 5b | OUD or opioid addiction under active treatment. |
| 6a | IV drug misuse with failed treatment or relapse. |
| 6b | IV drug misuse under active treatment. |
| 7a | Polysubstance misuse involving opioids. |
| 7b | Polysubstance misuse without opioids. |
| 8a | Confirmed nonstandard opioid administration. |
| 8b | Suspected nonstandard opioid administration. |
| 9 | Past or current medication treatment for OUD. |
| 10 | Naloxone administered before arrival, such as by EMS or in the field. |
| 11a | Confirmed opioid-seeking behavior. |
| 11b | Possible or nonspecific drug-seeking behavior. |

In the extraction stage, the LLM identified the evidence and provided a binary judgment (met/not met) for every item; if the judgment was 'met', the LLM returned the text-based evidence supporting the judgment. The extraction prompt directed the model to evaluate the items individually. The resulting 18-dimensional binary vector was a patient-level representation of the evidence found in the selected notes.

In the classification stage, a logistic regression layer uses the 18 indicators to determine a patient-level OUD probability. The fitted coefficients assigned a separate weight to each evidence item. The final output retained the item judgments, their supporting text, and the patient-level classification.

OPRO was incorporated into the proposed framework as the prompt-optimization stage. Within each outer training fold, OPRO refined the evidence-extraction prompt through three rounds, with four candidate prompts proposed per round. Candidate prompts could revise the system instruction, general extraction guidance, and evidence-item wording, while the 18 item identifiers and response schema remained fixed. We evaluated each candidate only on outer-training patients using 3-fold inner cross-validation of the complete evidence-extraction and patient-classification pipeline, with F1 as the optimization objective. We then froze the highest-performing prompt and applied it to the final performance evaluation.

**Experimental settings**

For machine learning-based CPs, we optimized hyperparameters separately for each representation and outer training fold using 3-fold inner cross-validation, with average precision as the selection criterion. All preprocessing, feature handling, model selection, and hyperparameter tuning were confined to the corresponding training data.

Zero-shot LLMs and evidence-extraction calls used a high reasoning profile, temperature zero, and a maximum completion length of 12,000 tokens. OPRO proposer calls used a high reasoning profile and temperature 0.8, whereas candidate evidence extraction used the same settings as the primary extraction calls. The context window was 128,000 tokens.

**Statistical analysis**

All approaches were evaluated using the chart-reviewed 253 patients using patient-level repeated stratified cross-validation splits, with 5 folds repeated 3 times and random seed 42, producing 15 shared outer test folds. Within each outer training set, we used preprocessing and 3-fold inner cross-validation to select the feature-based machine-learning learner and hyperparameters. For the OPRO-refined framework, prompt generation, candidate evaluation, and prompt selection were performed using only the corresponding outer-training patients. The selected prompt was then frozen, and the 18-item class-balanced logistic regression model was fitted using the outer-training data before the complete pipeline was applied to the corresponding outer-test patients. The Rubric-guided framework followed the same outer-fold aggregation and threshold-selection procedure without OPRO refinement. For each probabilistic approach, the threshold maximizing the Youden index in the outer-training data was applied unchanged to the corresponding outer-test fold. The diagnosis-code baseline used a fixed binary rule and required neither model fitting nor threshold selection.

We report sensitivity, specificity, positive predictive value (PPV), negative predictive value (NPV), F1, and area under the receiver operating characteristic curve (AUROC). We used average precision as the threshold-free, positive-class-oriented criterion for selecting machine-learning learners and hyperparameters. F1 was the primary operating-point metric. Values are fold means across the 15 outer test folds unless identified as unique-patient counts; between-method and prompt-refinement differences are summarized descriptively.

**Results**

Of the 300 patients chart-reviewed by physicians, 253 had definitive binary gold-standard labels and were included in the primary analysis, including 68 patients with OUD (26.9%) and 185 without OUD (73.1%). Demographic characteristics are presented in **Table 2**. Among the 40 patients independently reviewed by both physicians, agreement across positive, negative, and UNKNOWN assignments was 77.5%; among the 30 patients for whom both reviewers assigned definitive binary labels, agreement was 93.3% (28/30; Cohen's $\kappa = 0.865$). The remaining 47 patients were assigned UNKNOWN because chart review could not establish a definitive OUD-positive or OUD-negative classification and were excluded from binary evaluation.

**Table 2.** Demographic characteristics of the 253-patient binary evaluation cohort.

| Characteristic | Overall (N=253) | OUD positive (n=68) | OUD negative (n=185) |
|---|---|---|---|
| Age, median (IQR), years | 51 (40–58) | 41 (31–53) | 53 (44–61) |
| **Gender** | | | |
| Female | 161 (63.6%) | 49 (72.1%) | 112 (60.5%) |
| Male | 92 (36.4%) | 19 (27.9%) | 73 (39.5%) |
| **Race** | | | |
| White | 173 (68.4%) | 52 (76.5%) | 121 (65.4%) |
| Black or African American | 72 (28.5%) | 12 (17.6%) | 60 (32.4%) |
| Other or unknown | 8 (3.1%) | 4 (5.9%) | 4 (2.2 %) |

Values are n (%) unless otherwise indicated. Percentages were calculated within columns.

**Table 3** compares performance of all CP methods. Our OPRO-refined rubric-guided LLM achieved the highest F1 score (0.774) and AUROC (0.934), with a sensitivity of 0.858, specificity of 0.867, PPV of 0.715, and NPV of 0.944. Compared with the strongest feature-based model, EHR+NLP ML, the OPRO-refined approach yielded a higher F1 score (0.774 vs. 0.686) and AUROC (0.934 vs. 0.890). Relative to the zero-shot LLM, it demonstrated substantially higher specificity (0.867 vs. 0.602) and PPV (0.715 vs. 0.415), while maintaining higher sensitivity (0.858 vs. 0.775). Within the rubric-guided framework, OPRO refinement was associated with numerical improvements in F1 score (0.774 vs. 0.755) and AUROC (0.934 vs. 0.891), accompanied by smaller gains in sensitivity, specificity, PPV, and NPV.

**Table 3.** Comparison of OUD phenotyping approaches.

| Method | Sens. | Spec. | PPV | NPV | F1 | AUROC |
|---|---|---|---|---|---|---|
| Diagnosis code baseline | 0.662 | 0.719 | 0.466 | 0.854 | 0.545 | — |
| EHR-only ML | 0.667 | 0.822 | 0.603 | 0.873 | 0.622 | 0.830 |
| NLP-only ML | 0.671 | 0.843 | 0.626 | 0.876 | 0.641 | 0.829 |
| EHR+NLP ML | 0.770 | 0.823 | 0.626 | 0.908 | 0.686 | 0.890 |
| Zero-shot LLM | 0.775 | 0.602 | 0.415 | 0.887 | 0.536 | 0.805 |
| Rubric-guided LLM | 0.832 | 0.861 | 0.700 | 0.935 | 0.755 | 0.891 |
| OPRO-refined rubric-guided LLM | **0.858** | **0.867** | **0.715** | **0.944** | **0.774** | **0.934** |

All LLM-based approaches used gpt-oss-120b

The OPRO module optimizes the system prompt, general extraction instructions, and operational definitions of four evidence items while preserving the item identifiers and fixed response schema. The OPRO-selected prompt reframed the system role from OUD adjudication to explicit evidence extraction. It also introduced broadened definitions for items 4a, 5b, 9, and 10, while retaining the original criteria for all remaining items. Table 4 summarizes **these refinements**.

**Table 4.** Comparison between seed prompt and OPRO-refined prompt.

| Prompt component | Seed prompt | OPRO-refined prompt |
|---|---|---|
| System prompt | You are a meticulous clinical chart reviewer specializing in OUD adjudication. | You are a clinical informatics specialist tasked with extracting exact evidence for OUD from patient notes. |
| General extraction instructions | Evaluate each rule independently and literally against its definition. | Use the following broadened definitions when scanning the notes. For all other items, retain the original literal criteria and provide a verbatim quotation for every item marked as present. |
| Item 4a: confirmed opioid misuse | Documented confirmed history of opioid misuse or abuse | Added explicit references to opioid dependence, OUD, F11.x codes, heroin or fentanyl use, and buprenorphine treatment for OUD |
| Item 5b: OUD under active treatment | Opioid addiction with a current treatment plan | Specified ongoing methadone, buprenorphine/Suboxone, naltrexone, counseling, or rehabilitation |
| Item 9: medication treatment for OUD | History of medication-assisted or other opioid addiction treatment | Added past or present OUD medication and the terms MAT, opioid agonist therapy, and opioid antagonist therapy |
| Item 10: naloxone before arrival | Narcan or naloxone administered before intake | Added EMS, field, and prehospital documentation of naloxone administration |

## Discussion

Identifying OUD cases is a challenging, complex task that requires supporting evidence dispersed across clinical narratives and expressed using heterogeneous terminologies[7,17]. This study explored state-of-the-art LLMs to improve clinical CP of OUD using clinical text and structured EHRs[18]. The experimental results show that our LLM-based solution achieved the best CP performance, outperforming traditional machine learning models and zero-shot LLM baselines. This study demonstrates the advantage of using rule-guided LLMs for clinical phenotyping of OUD.

The proposed LLM solution formulated OUD CP as two tasks: identifying prespecified evidence in clinical narratives and using the extracted evidence to determine CP. The LLM first evaluated 18 evidence items and returned supporting text evidence. These evidence items distinguished confirmed, suspected, historical, treatment-related, and nonspecific OUD-related observations. A logistic regression layer then uses LLM-identified evidence to determine the OUD flag and provide text evidence. The proposed LLM-based CP uses both the evidence extracted from the narratives and the patient-level information from structured EHRs. We use the OPRO algorithm to refine the evidence-extraction prompt while retaining the same 18 item identifiers, response schema, and patient-classification procedure. We observed improved performance with OPRO-optimized prompts compared with the seed prompt, indicating the efficiency of OPRO optimization. Prompt generation and selection were restricted to the training data in each outer fold before applying the selected prompt to the corresponding test patients. As an LLM-based solution, our LLM-based CP could provide text-based evidence and link it to the final CP judgment. The experimental results show that OPRO-refined extraction and evidence linking could provide an LLM-based CP for evidence-based OUD phenotyping.

This study has limitations. The framework was evaluated in a single site, and its performance at other healthcare systems may differ in populations with different OUD prevalence and documentation practices. Like all EHR-based phenotypes, the framework can identify only evidence documented in the available clinical record. External validation is therefore needed across institutions and patient populations.

**Conclusion**

This study developed a rubric-guided LLM for OUD computable phenotyping that is based on gpt-oss-120b. Our LLM-based solution achieved the highest fold-mean F1 score (0.774) and AUROC (0.934) among the evaluated approaches. Our LLM-based CP improved OUD case identification and provides a traceable, evidence-based approach to retrospective OUD cohort identification.

**Acknowledgment**

This study was partially supported by grants from the Patient-Centered Outcomes Research Institute® (PCORI®) Award ME-2023C3-35934, the PARADIGM program awarded by the Advanced Research Projects Agency for Health (ARPA-H), National Institute on Aging U24AG098157, National Institute of Allergy and Infectious Diseases, NIAID R01AI172875, National Heart, Lung, and Blood Institute, R01HL169277, R01HL176844, National Institute on Drug Abuse, NIDA R01DA057886, R01DA063631, and the UF Clinical and Translational Science Institute. The content is solely the responsibility of the authors and does not necessarily represent the official views of the funding institutions.